\def\year{2022}\relax
\documentclass[letterpaper]{article} 
\usepackage{aaai22}  
\usepackage{times}  
\usepackage{helvet}  
\usepackage{courier}  
\usepackage[hyphens]{url}  
\usepackage{graphicx} 
\usepackage{natbib}  
\usepackage{caption} 
\DeclareCaptionStyle{ruled}{labelfont=normalfont,labelsep=colon,strut=off} 
\usepackage{algorithm}
\usepackage{algorithmic}

\usepackage{newfloat}
\usepackage{listings}
\floatstyle{ruled}
\newfloat{listing}{tb}{lst}{}
\floatname{listing}{Listing}
\title{Zero-Shot Cross-Lingual Machine Reading Comprehension\\ via Inter-sentence Dependency Graph}
\author{
Liyan Xu\textsuperscript{\rm 1},
Xuchao Zhang\textsuperscript{\rm 2},
Bo Zong\textsuperscript{\rm 2},
Yanchi Liu\textsuperscript{\rm 2},
Wei Cheng\textsuperscript{\rm 2},\\
Jingchao Ni\textsuperscript{\rm 2},
Haifeng Chen\textsuperscript{\rm 2},
Liang Zhao\textsuperscript{\rm 1},
Jinho D. Choi\textsuperscript{\rm 1}
}
\affiliations {
\textsuperscript{\rm 1}Department of Computer Science, Emory University, Atlanta, GA, USA\\
\textsuperscript{\rm 2}NEC Laboratories America, Princeton, NJ, USA\\
\textsuperscript{\rm 1}\{liyan.xu, liang.zhao, jinho.choi\}@emory.edu \\
\textsuperscript{\rm 2}\{xuczhang, bozong, yanchi, weicheng, jni, haifeng\}@nec-labs.com\\
}

\usepackage{enumitem}
\usepackage{amsfonts,amsmath,amssymb}
\usepackage{array,booktabs}
\newcommand{\RN}[1]{%
  \textup{\uppercase\expandafter{\romannumeral#1}}%
}

\begin{document}

\maketitle

\begin{abstract}
We target the task of cross-lingual Machine Reading Comprehension (MRC) in the direct zero-shot setting, by incorporating syntactic features from Universal Dependencies (UD), and the key features we use are the syntactic relations within each sentence.
While previous work has demonstrated effective syntax-guided MRC models, we propose to adopt the inter-sentence syntactic relations, in addition to the rudimentary intra-sentence relations, to further utilize the syntactic dependencies in the multi-sentence input of the MRC task.
In our approach, we build the Inter-Sentence Dependency Graph (ISDG) connecting dependency trees to form global syntactic relations across sentences.
We then propose the ISDG encoder that encodes the global dependency graph, addressing the inter-sentence relations via both one-hop and multi-hop dependency paths explicitly.
Experiments on three multilingual MRC datasets (XQuAD, MLQA, TyDiQA-GoldP) show that our encoder that is only trained on English is able to improve the zero-shot performance on all 14 test sets covering 8 languages, with up to 3.8 F1 / 5.2 EM improvement on-average, and 5.2 F1 / 11.2 EM on certain languages.
Further analysis shows the improvement can be attributed to the attention on the cross-linguistically consistent syntactic path.
Our code is available at \url{https://github.com/lxucs/multilingual-mrc-isdg}.
\end{abstract}

\section{Introduction}
\label{sec:introduction}

Universal Dependencies (UD) \citep{nivre-etal-2016-universal} is a unified framework that aims to provide cross-linguistically consistent features including part-of-speech (POS) tags, morphological features and syntactic dependencies for over 90 languages. With the recent release of more than 100 treebanks thanks to great annotation efforts
, several toolkits have been made available, such as Stanza \citep{qi-etal-2020-stanza} and UDPipe \citep{straka-2018-udpipe}, which are built upon the UD framework and provide state-of-the-art performance on predicting universal syntactic features for multiple languages, offering new potentials for cross-lingual applications.

\begin{figure}[t]
\centering
\includegraphics[width=0.84\columnwidth]{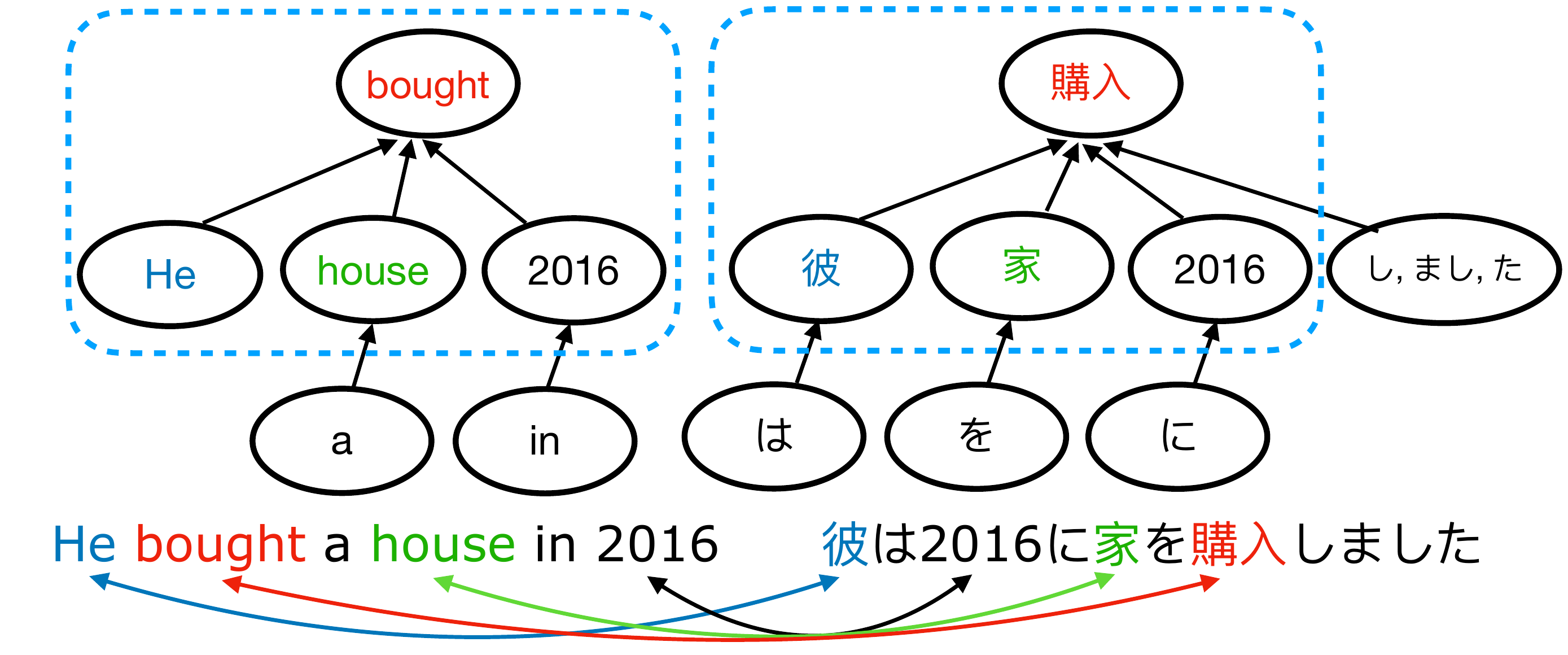}
\caption{Syntactic dependency representation of parallel sentences in English and Japanese. We mark the aligned verbs and nouns of the same meaning by the same color. Two languages have quite different sentence structure, while the main components (verbs and nouns) have the same graph structure under syntactic dependencies, reducing the cross-lingual gap on the representation.}
\label{fig:motivation}
\end{figure}

In this work, we target to incorporate UD features in the task of zero-shot cross-lingual machine reading comprehension (MRC), exploiting the potentials brought by UD.
Specifically, our main motivation is that the raw text of each language can exhibit its own unique linguistic traits, while the cross-linguistically consistent syntax can serve as the anchor points across multiple languages. For example, Figure~\ref{fig:motivation} shows the parallel sentences in English and Japanese that vary quite a lot in sentence structure. By providing the extra clues of universal syntactic dependencies, the model can benefit from  a closer gap of cross-lingual representation with the explicit alignment from the dependency graph structure.

Various past work has already shown syntactically informed models to be effective in machine translation \citep{AAAI1816060, zhang-etal-2019-syntax} and other monolingual intra-sentence tasks such as Semantic Role Labeling (SRL) \citep{strubell-etal-2018-linguistically, kasai-etal-2019-syntax}.
Recently, the utilization of additional syntactic clues seems to depreciate, as the emerging pretrained language models such as BERT \citep{devlin-etal-2019-bert} already implicitly encode the linguistic notions of syntax \cite{hewitt-manning-2019-structural}.
Nevertheless, the values of this work are twofold.
First, existing methods focus on the direct syntactic relations within each sentence, e.g.  the recent MRC model SG-Net \cite{zhang_sg-net_2020}; while we further explicitly address the multi-hop relations in a global syntactic graph across sentences.
Second, whether the syntactic features can provide useful auxiliary information for multilingual MRC is still an open question that has not been answered before, given the fact that the universal syntax is just made available recently.

Our approach adopts the multilingual pretrained language models as the backbone, and features the direct zero-shot transfer, where the entire model is trained only on the source language and evaluated directly on the test sets in multiple target languages.
Our proposed model aims to be an augmentation upon any pretrained models, and can be further combined with other cross-lingual transfer techniques that involve target languages in the training, such as adding translation to target languages in the training \citep{hsu-etal-2019-zero,lee-etal-2019-learning,cui-etal-2019-cross, yuan-etal-2020-enhancing}.

To address the major challenge of utilizing syntactic dependencies in the multi-sentence documents of the MRC task, we first build the Inter-Sentence Dependency Graph (ISDG), which is a document-level graph that connects the syntactic dependencies of each sentence (Section~\ref{subsec:ud-features}).
We then introduce our ISDG encoder stacked upon the pretrained language model, which is a graph encoder based on self-attention \citep{self-attention} and specifically encodes the ISDG structure and relations.
The proposed encoder consists of two components: the ``local'' component that models the local one-hop relations directly among graph nodes; the ``global'' component that focuses on the global multi-hop relations, and explicitly models the syntactic dependencies across sentences.
In particular, we define ``soft'' paths that approximate the full paths between every node pair, based on the unique characteristic of ISDG, and inject the paths as the new representation of keys and queries in self-attention.

We conduct experiments with three different pretrained language models on three multilingual MRC datasets to test the generalizability of our approach: XQuAD \citep{artetxe-etal-2020-cross}, MLQA \citep{lewis-etal-2020-mlqa}, TyDiQA-GoldP \citep{clark-etal-2020-tydi}.
The evaluation covers 14 test sets in 8 languages that are supported by UD. Empirical results show that our proposed graph encoder is able to improve the zero-shot performance on all test sets in terms of either F1 or EM, boosting the on-average performance on all three datasets by up to 3.8 F1 and 5.2 EM (Section~\ref{subsec:results}), and obtains up to 5.2 F1 / 11.2 EM improvement on certain languages.
Results suggest that the zero-shot model is able to benefit from the cross-linguistically consistent UD features for most experimented languages, and the analysis shows that the proposed attention on the global inter-sentence syntactic dependencies could play an important role.

\section{Related Work}
\label{sec:related-work}

We categorize zero-shot cross-lingual transfer (CLT) into two types. The first type is the direct transfer, where the training only involves the source language without exposing any target languages. Recent multilingual pretrained language models have brought significant advances to the direct transfer performance by aligning different languages to the shared embedding space, such as mBERT \citep{devlin-etal-2019-bert},
XLM-R \citep{conneau-etal-2020-unsupervised}, mT5 \citep{xue2020mt5}.
The second type of zero-shot CLT is to expose certain target languages directly in the training process, and many techniques have been proposed within this line of work. In the task of MRC, \citet{hsu-etal-2019-zero, lee-etal-2019-learning, cui-etal-2019-cross} obtain training corpus for target languages by utilizing translation and projecting silver labels; similar techniques are also used in other cross-lingual tasks such as SRL \citep{cai-lapata-2020-alignment}, POS tagging \citep{eskander-etal-2020-unsupervised} and Abstract Meaning Representation (AMR) parsing \citep{blloshmi-etal-2020-xl}. Other techniques such as self-learning \citep{xu-etal-2021-boosting} and meta-learning \citep{li-etal-2020-learn, nooralahzadeh-etal-2020-zero} are also proposed for CLT.
Our work is an augmentation of the first CLT type; however, it does not conflict with the second type, and can be further combined with other techniques that involve target languages in the training.

Previous work has introduced various syntax-guided graph models mostly under the monolingual setting.
Early work includes Tree-LSTM \citep{tai-etal-2015-improved} and Graph-LSTM \citep{song-etal-2018-graph} to encode syntactic trees or AMR graphs.
Several recent work on the AMR-to-text task \citep{ guo-etal-2019-densely,subburathinam-etal-2019-cross}
uses variants of Graph Convolutional Network (GCN)
\citep{Kipf:2016tc}
in the graph encoding. Our proposed encoder is closer to some other recent work \citep{zhu-etal-2019-modeling, cai_graph_2020, yao-etal-2020-heterogeneous, zhang_sg-net_2020} that encodes graphs in self-attention.
Our approach is distinguished from previous work as we address both the zero-shot multilingual perspective as well as the global dependencies in the multi-sentence input.

\section{Approach}
\label{sec:approach}

We first briefly review the multilingual pretrained language model, which is the baseline and used as the backbone in our experiments.
We then introduce features from UD, and how we encode the syntactic features using both local and global encoding components in our proposed ISDG encoder.

\subsection{Multilingual Pretrained Models}
\label{subsec:multilingual-encoder}

Recent multilingual pretrained language models adopt the Transformers architecture \citep{self-attention} for sequence encoding, and their direct zero-shot performance is used as the baseline.
Following the previous work on the span-extraction MRC task, we use the same input format where the question and context are packed in a single sequence. We also use the same decoding scheme in all our experiments, where two linear layers are stacked on the encoder to predict the start and end positions of the answer span respectively. The log-likelihoods of the gold start and end positions $i_s, i_e$ are being optimized during training:
\begin{align}
    p^{s/e}(i) &= \text{softmax} \big( W^{s/e}_L x_i + b^{s/e}_L \big)\\
    \mathcal{L} &= -\log p^s(i_s) - \log p^e(i_e)
\end{align}
where $p^{s/e}(i)$ is the likelihood of token $i$ being the start/end position, $W^{s/e}_L$ and $b^{s/e}_L$ are the parameters for the linear layers, and $\mathcal{L}$ is the loss function. The final selected prediction is the span with the highest sum of start and end likelihood.

\subsection{Universal Dependencies}
\label{subsec:ud-features}

\begin{figure*}[tp]
\centering
\includegraphics[width=0.95\textwidth]{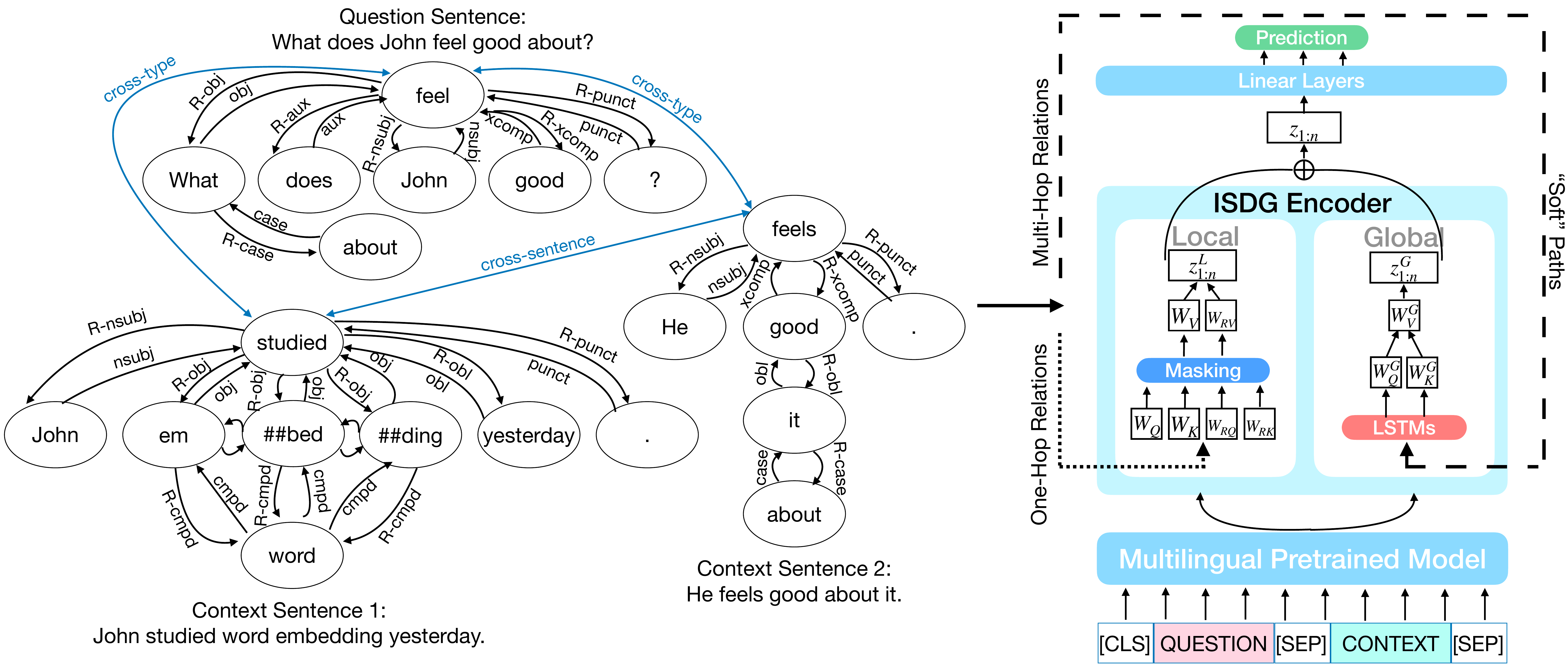}
\caption{On the left side, a simplified example of the ISDG is shown. Nodes are connected by syntactic dependency relations; reverse relations are prepended by ``R-''. Special types of \textit{cross-sentence} and \textit{cross-type} connect root nodes of the dependency trees, marked by the blue color. For simplicity, we omit the self-connection on each node, and omit the \textit{subtoken} relations among subtokens of ``em'', ``\#\#bed'', ``\#\#ding''. On the right side, an overview of our model architecture is shown. Our proposed ISDG encoder is stacked 
upon the pretrained language model, and encodes the local one-hop and global multi-hop dependency relations in the obtained multi-sentence graph structure.}
\label{fig:main}
\end{figure*}

\begin{figure}[t]
\centering
\includegraphics[width=0.8\columnwidth]{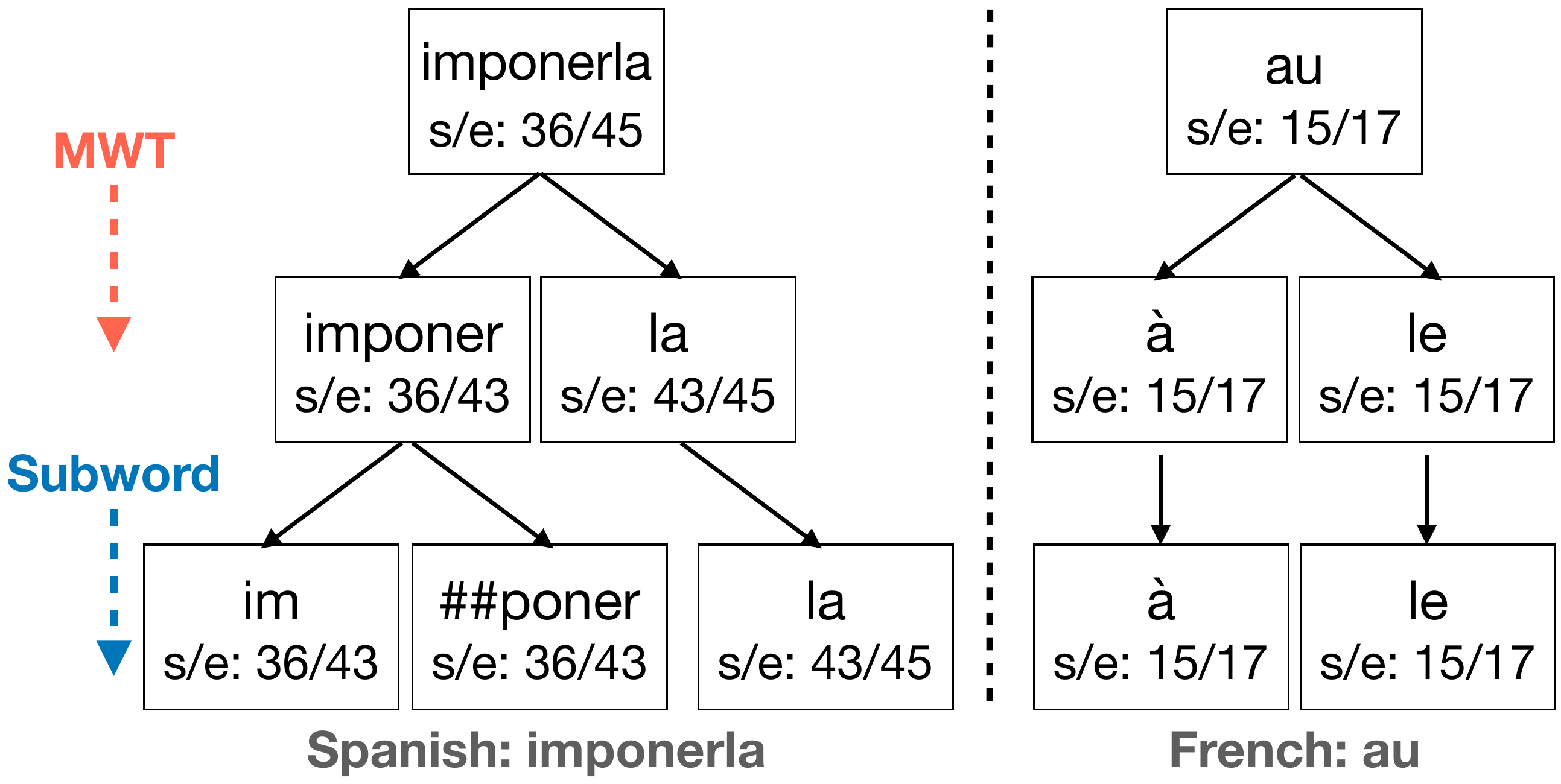}
\caption{Examples of the tokenization process. s/e represents the index of the start/end character in the raw text.}
\label{fig:tokenization}
\end{figure}

\paragraph{UD Tokenization}
Since all raw UD features are based on UD's own tokenization, we first adapt our model to accommodate the tokenization from both UD and the pretrained model. Specifically, UD first tokenizes the raw text into raw tokens, and then applies the ``Multi-Word Token (MWT) expansion'' on each token, which could change its morphological form and further split off multiple words, and each word can have completely different text that does not appear in the original text. We address this by building a heuristic mapping from each word (after MWT expansion) to its start and end character index in the original text, and then perform the tokenization of the pretrained model on each word to obtain the subtokens, as shown in Figure~\ref{fig:tokenization}.

The left side of Figure~\ref{fig:tokenization} shows an example in Spanish where MWT simply splits ``imponerla'' into two words by adding segmentation; in this case, we can obtain the indices of the start and end characters of the resulting words accordingly. The right side shows an example in French where MWT splits ``au'' into two words of different text. In this case, we assign their character indices to be the same as the original token, since the words after MWT do not exist in the raw text.
To generate the predicted answer, we can then simply use the leftmost and rightmost character index of the predicted subword position to recover the text span.




\paragraph{Universal POS}
We use a learnable embedding layer for the 17 POS types defined by UD. For each subtoken, we concatenate its POS embedding along with its hidden state from the last layer of the pretrained models, serving as the new input hidden state for the following graph encoder.

\paragraph{Universal Syntactic Dependencies}

UD provides the syntactic dependency features for each word (after MWT expansion) in a sentence, including its head word and the dependency relation to the head word.
Each sentence contains one unique root word with no head word.
In this work, we use the main relation types from UD, without considering subtypes.
The syntactic dependency features are consumed by the proposed model as follows.

\subsection{Inter-Sentence Dependency Graph (ISDG)}
\label{subsec:isdg}
Since MRC is a document-level task, the input usually consists of multiple sentences for the context and question.
While previous work has focused on encoding the raw syntactic dependencies within each sentence directly, we propose to further consider the global syntactic relations that strengthen the document-level input structure.
Therefore, we build the following graph for the multi-sentence input, using the dependency trees of each sentence to build global syntactic relations, namely Inter-Sentence Dependency Graph (ISDG). An example of ISDG is shown in Figure~\ref{fig:main}.

We first obtain the original dependency tree of each sentence, and also add the reserve relation from each head word to its child words. We then adapt the tree to the subtoken level: we split each word into nodes of its corresponding subtokens, where each subtoken node shares the same relations as the word. Among all subtokens from the same word, we fully connect them by a special relation \textit{subtoken}, and also self-connect each node by a special relation \textit{self}. For special subtokens such as \texttt{[CLS]} and \texttt{[SEP]}, only the self-connections are assigned. All ``nodes'' in the rest of this paper refer to the graph nodes on the subtoken level.

We then connect all the independent dependency trees to construct the final ISDG. Specifically, we fully connect all the root nodes within the context sentences with a special relation \textit{cross-sentence}, and use another special relation \textit{cross-type} to fully connect all root nodes between the question and context sentences, to distinguish the dual input types. Thus, each node in ISDG can reach to any other node through a one-hop or multi-hop dependency path, building the global syntactic relations. The design objective of ISDG is to keep all raw syntactic features as well as adding the visibility of the cross-sentence input structure.

\subsection{ISDG Encoder: Local Encoding}
\label{subsec:graph-encoder-local}

For each input, our proposed ISDG encoder is dedicated to encode its ISDG obtained above, and it consists of two components: the local encoding component that focuses on the local one-hop relations directly (Section~\ref{subsec:graph-encoder-local}), and the global encoding component that further accounts for the global multi-hop syntactic relations across sentences (Section~\ref{subsec:graph-encoder-global}).

The local encoding component adapts the idea of relative position encoding that has been explored by several recent work \citep{shaw-etal-2018-self,dai-etal-2019-transformer,cai_graph_2020}.
We denote the hidden state of each input node at sequence position $i$ as $x_i$, which is the concatenation of its POS embedding and its hidden state from the pretrained model.
The hidden state of the relation type from node $i$ to node $j$ is denoted as $r_{ij}$, which is obtained from a separate learnable embedding layer.
The structure of one-hop relations are injected into the self-attention as follows:
\begin{align}
\label{eq:vanilla-graph}
    e_{ij}^{\text{L}} &= \big( (x_i + r_{ij}) W_Q \big) \big( (x_j + r_{ji}) W_K \big)^T\\[5pt]
    &= \underbrace{(x_i W_Q W_K^T x_j)}_{(a)} + \underbrace{(x_i W_Q W_K^T r_{ji})}_{(b)} \nonumber\\
    &+ \underbrace{(r_{ij} W_Q W_K^T x_j)}_{(c)} + \underbrace{(r_{ij} W_Q W_K^T r_{ji})}_{(d)} \nonumber
\end{align}
$e_{ij}^{\text{L}}$ is the raw attention score that takes into account the local one-hop relation type from node $i$ to $j$ in ISDG; $W_Q$ and $W_K$ are the query and key parameters.
In particular, Eq~\eqref{eq:vanilla-graph} can be decomposed and interpreted by four parts.
The term (a) is the same as the original self-attention; the term (b) and (c) represent the relation bias conditioned on the source/target node; the term (d) is the prior bias on the relation types.

However, the vanilla injection in Eq~\eqref{eq:vanilla-graph} cannot fit for ISDG directly, and we make two adaptations to address the following issues.

First, let $d_x$ and $d_r$ be the hidden size of nodes and relations; Eq~\eqref{eq:vanilla-graph} requires equal hidden sizes $d_x = d_r$.
For each input sequence, the embedding matrices of nodes and relations have sizes $n d_x$ and $n^2 d_r$ respectively. Therefore, it would be impractical to keep $d_x = d_r$ for the document-level task  where $n$ can be quite large.
We make the first adaptation that sets $d_r$ to be much smaller than $d_x$ and uses another set of key and query parameters for the relations. We also share the relation matrix across attention heads to reduce the memory usage.

Second, since ISDG is not a complete graph, we implicitly set a \textit{none} type for any $r_{ij}$ with no relations. However, this would introduce a non-trivial inductive bias in Eq~\eqref{eq:vanilla-graph},
as \textit{none} type can be prevalent in the graph matrix. Thus, we apply attention masking $\mathcal{M}$ on the attention scores by the \textit{none} type specified in Eq~\eqref{eq:masking1} and \eqref{eq:masking2}, similar to \citet{yao-etal-2020-heterogeneous, zhang_sg-net_2020},
enforcing the inductive bias to be 0 among nodes that are not directly connected.

Lastly, we also inject the relations into the value representation of self-attention as in Eq~\eqref{eq:local-final}. The final normalized attention score $\alpha^{\text{L}}$ and output $z^{\text{L}}$ are computed as:
\begin{align}
\mathcal{M}_{ij} &= 
    \begin{cases}
        1 \quad r_{ij} \neq \textit{none}\\
        0 \quad \text{otherwise}
    \end{cases} \label{eq:masking1} \\
    \alpha_{ij}^{\text{L}} &= \frac{\exp (\mathcal{M}_{ij} \cdot e_{ij}^{\text{L}} / \sqrt{d_x})}{\sum_{k=1}^n \exp (\mathcal{M}_{ik} \cdot e_{ik}^{\text{L}} / \sqrt{d_x})} \label{eq:masking2} \\
    z_i^{\text{L}} &= \sum_{j=1}^n \alpha_{ij}^{\text{L}} (x_j W_V + r_{ij} W_{RV}) \label{eq:local-final}
\end{align}
$W_{V} \in \mathbb{R}^{d_x \times d_x}$ and $W_{RV} \in \mathbb{R}^{d_r \times d_x}$ are the query parameters for the nodes and relations.
Note that multiple layers of the local encoding component can be stacked together to implicitly model the higher-order dependencies, however in practice, stacking multiple layers are constrained by the GPU memory, and quickly becomes impractical under the huge document-level graph matrix.

\subsection{ISDG Encoder: Global Encoding}
\label{subsec:graph-encoder-global}

We next propose and integrate the following global encoding component into the ISDG encoder, for the fact that each pair of nodes in ISDG always has a dependency path of relations, and making use of this multi-hop relations should further provide stronger sequence encoding.
Previous work has addressed multi-hop relations by directly encoding the shortest path between two nodes for sentence-level tasks \citep{zhu-etal-2019-modeling, cai_graph_2020}.
However, this is not practical for the MRC task, as the sequence length $n$ can be much larger for the document-level input.
Let $l_p$ be the maximum path length, $d_p$ be the hidden size for each path step. The size of the path matrix is $n^2 l_p d_p$ that includes each pair of nodes, which can easily consume all GPU memory.

To address the above challenge, our proposed global encoding component utilizes an approximated path between any two nodes, rather than the full path.
We refer to it as the ``soft'' path, which has a much lower space complexity than the full path matrix, making it possible for the model to encode the multi-hop relations give the long input sequence.

\begin{figure}[ht]
\centering
\includegraphics[width=0.8\columnwidth]{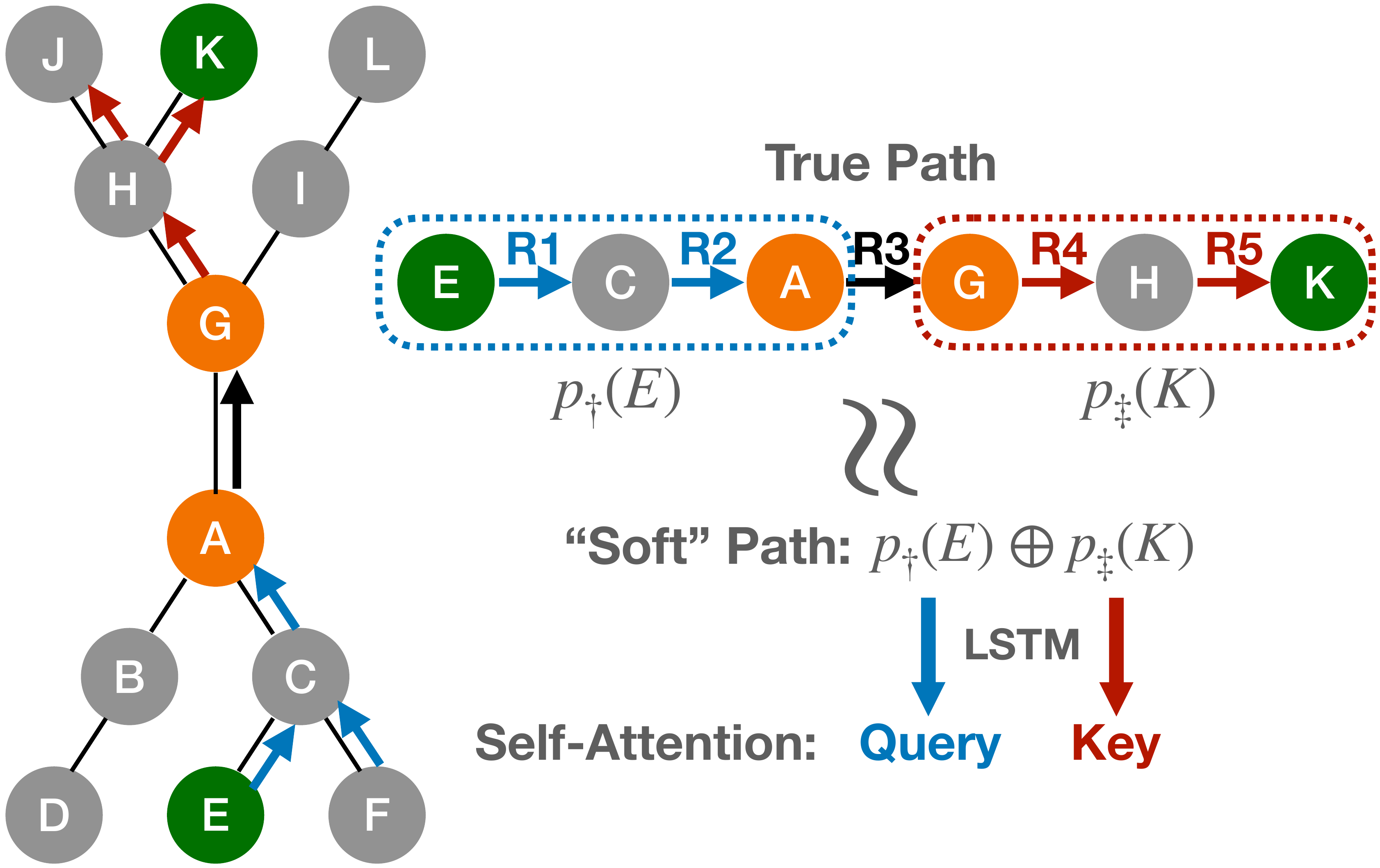}
\caption{Illustration of the ``soft'' path. Two dependency trees are depicted with root nodes A and G. True paths of all node pairs are heavily overlapped, as each node needs to go through its root node. The ``soft'' path from node E to K is shown, which is the concatenation of the outgoing path of node E:  $p_{\dagger}(E)$, and the incoming path of node K:  $p_{\ddagger}(K)$, as an approximation of the true path.}
\label{fig:soft}
\end{figure}

The rationale behind ``soft'' paths is the observation that the paths of many node pairs are heavily overlapped: for any cross-sentence node pairs, each of the node always goes through its root node. We denote $p_{\dagger}(i)$ as the outgoing path of hidden states from node $i$ to its root node $i_r$:
\begin{align}
    p_{\dagger}(i) = (x_i, r_{ik_1}, x_{k_1}, r_{k_1 k_2}, \dots, r_{k_{i} i_r}, x_{i_r})\nonumber
\end{align}
with $k_1, \dots, k_i$ being the intermediate nodes in the path. Similarly, we denote $p_{\ddagger}(i)$ as the incoming path from root node $i_r$ to node $i$, which has the reverse order of $p_{\dagger}(i)$. We then define the ``soft'' path $\tau_{ij}$ from node $i$ to $j$ as:
\begin{align}
    \tau_{ij} &= (x_i, \dots, x_{ir}, x_{jr}, \dots, x_j)\nonumber\\
            &= p_{\dagger}(i) \oplus p_{\ddagger}(j)
\end{align}
$x_{ir}$ and $x_{jr}$ are the root nodes for $i$ and $j$, $\oplus$ denotes the concatenation. $\tau_{ij}$ largely captures the true shortest paths of cross-sentence node pairs and only loses one intermediate relation $r_{i_r j_r}$ between the two root nodes; for within-sentence pairs, $\tau_{ij}$ can become non-shortest path, but still provides auxiliary information over the direct one-hop relations in the local encoding component. An illustration of the ``soft'' paths are shown in Figure~\ref{fig:soft}.

As a result of the ``soft'' path trade-off, we can now fit the approximated path of global multi-hop relations into self-attention.
We encode the outgoing and incoming ``soft'' paths by long short-term memory (LSTM), of which hidden states are denoted by $\overrightarrow{h_{i,t}}$ and $\overleftarrow{h_{i, t}}$ at the step $t$ for the node $i$:
\begin{align}
    \overrightarrow{h_{i,t}} &\leftarrow \text{LSTM}(s_{i,t}^{\dagger}, \overrightarrow{h_{i, t-1}}; \theta^{\dagger})\\
    \overleftarrow{h_{i,t}} &\leftarrow \text{LSTM}(s_{i,t}^{\ddagger}, \overleftarrow{h_{i, t-1}}; \theta^{\ddagger})
\end{align}
where $s_{i,t}^{\dagger}$ and $s_{i,t}^{\ddagger}$ are the \textit{t}th hidden states in the ``soft'' path $p_{\dagger}(i)$ and $p_{\ddagger}(i)$; $\theta^{\dagger}$ and $\theta^{\ddagger}$ are the parameters for LSTMs.

We can then obtain two distinct representation for each node $i$, denoted by $\overrightarrow{g_i}$ and $\overleftarrow{g_i}$, which are the last LSTM hidden states of the outgoing path $p_{\dagger}(i)$ and incoming path $p_{\ddagger}(i)$ respectively. 
We make the outgoing path representation $\overrightarrow{g_i}$ of node $i$ as the query, and make incoming path representation $\overleftarrow{g_j}$ of node $j$ as the key, resembling the ``soft'' path $\tau_{ij}$ to be injected into the self-attention:
\begin{align}
\label{eq:sp_sa}
    e_{ij}^{\text{G}} &= (\overrightarrow{g_i} W_Q^{\text{G}})(\overleftarrow{g_j} W_K^{\text{G}})^T\\
    \alpha_{ij}^{\text{G}} &= \frac{\exp (e_{ij}^{\text{G}} / \sqrt{d_x})}{\sum_{k=1}^n \exp (e_{ik}^{\text{G}} / \sqrt{d_x})}\\
    z_i^{\text{G}} &= \sum_{j=1}^n \alpha_{ij}^{\text{G}} \big( (\overrightarrow{g_i} + \overleftarrow{g_j}) W_V^{\text{G}} \big)\\
    z_i &= z_i^{\text{L}} \oplus z_i^{\text{G}}
\end{align}
$W_Q^{\text{G}}, W_K^{\text{G}}, W_V^{\text{G}} \in \mathbb{R}^{d_x \times d_x}$ are the query, key, value parameters for the global encoding component. The final output of the ISDG encoder $z_i$ is the concatenation of the output from both local and global encoding components. To further strengthen inter-sentence interaction, additional layers of vanilla self-attention can optionally be stacked upon the ISDG encoder that takes the output sequence $z_{1:n}$ as input.

\section{Experiments}
\label{sec:experiments}

\subsection{Implementation Settings}
\label{subsec:impl}

We implement our models in PyTorch and use Stanza \citep{qi-etal-2020-stanza} to provide the UD features.
Obtaining UD features for training and test sets is computed as an offline preprocessing step, taking around 4 hours.

We experiment with three recent multilingual pretrained language models: mBERT \citep{devlin-etal-2019-bert}, XLM-R\textsubscript{Large} \citep{conneau-etal-2020-unsupervised}, mT5\textsubscript{Large} \citep{xue2020mt5}. For fair comparison, we maintain the following conditions identical per the pretrained model and per the dataset: (1) same pretrained weights and hyperparameters; (2) same decoding scheme (Section~\ref{subsec:multilingual-encoder}). For mBERT and XLM-R\textsubscript{Large}, we follow the similar hyperparameter settings as XTREME, with 384 max sequence length and 2 training epochs. For mT5\textsubscript{Large}, we only use its encoder and discard the decoder, and employ a learning rate of $1 \times 10^{-4}$, which achieves the same baseline results as reported by \citet{xue2020mt5}.




For experiments with ISDG, we limit the max path length to be 8, and truncate long ``soft'' paths from the end. 64 hidden size is adopted for the POS and relation embedding. Following SG-Net \citep{zhang_sg-net_2020}, we append one final self-attention layer stacked upon the ISDG encoder. All experiments are conducted on a Nvidia A100 GPU, with training time around 1 - 2 hours for the baseline and 2.5 - 4 hours for the ISDG encoder.

\subsection{Evaluation Protocols}
\label{subsec:eval}

\begin{table*}[t!]
\centering
\resizebox{0.72\textwidth}{!}{
\begin{tabular}{l|cccccc|c|}
& en & de & el & es & hi & ru & \bf avg \\
\midrule
mBERT\textsuperscript{*} & 83.5 / 72.2 & 70.6 / 54.0 & 62.6 / 44.9 & 75.5 / 56.9 & 59.2 / \textbf{46.0} & 71.3 / 53.3 & 70.5 / 54.6 \\
mBERT & 83.8 / 73.0 & 71.7 / 55.8 & 63.6 / 45.8 & \bf 76.4 / 59.0 & 58.2 / 44.0 & 71.5 / 55.1 & 70.9 / 55.5 \\
\; + ISDG & \bf 84.1 / 73.1 & \bf 74.1 / 57.6 & \bf 64.4 / 48.2 & 76.1 / 57.8 & \bf 59.3 / 46.0 & \bf 72.2 / 55.3 & \bf 71.7 / 56.3 \\
\midrule
XLM-R\textsuperscript{*} & 86.5 / 75.7 & 80.4 / 63.4 & 79.8 / 61.7 & 82.0 / 63.9 & 76.7 / 59.7 & 80.1 / 64.3 & 80.9 / 64.8 \\
XLM-R & 87.4 / 76.3 & 80.8 / 63.9 & 80.6 / 63.4 & 82.2 / 63.0 & 76.4 / 60.0 & 80.9 / \textbf{\underline{65.1}} & 81.4 / 65.3 \\
\; + ISDG & \bf 88.6 / 77.9 & \textbf{82.1} / \textbf{\underline{66.1}} & \bf \underline{81.9} / \underline{64.3} & \bf \underline{83.4} / \underline{65.9} & \bf \underline{76.9} / \underline{60.9} & \textbf{\underline{81.3}} / 64.5 & \bf \underline{82.4} / \underline{66.6} \\
\midrule
mT5\textsuperscript{*} & 88.4 / 77.3 & 80.0 / 62.9 & 77.5 / 57.6 &  81.8 / \textbf{64.2} & 73.4 / 56.6 & 74.7 / 56.9 & 79.3 / 62.6 \\
mT5 & 87.8 / 76.8 & 80.9 / 63.9 & 79.3 / 60.9 & \textbf{82.4} / 64.0 & 75.7 / 58.7 & 78.6 / 62.2 & 80.8 / 64.4 \\
\; + ISDG & \bf \underline{88.7} / \underline{78.2} & \textbf{\underline{82.5}} / \textbf{65.4} & \bf 80.5 / 61.3 & 82.1 / 63.2 & \textbf{\underline{76.9}} / \textbf{60.3} & \bf 80.5 / 64.2 & \bf 81.9 / 65.4 \\
\end{tabular}}
\caption{XQuAD results (F1/EM) for each language. * denotes the results from original papers. Bold numbers are the best results per pretrained language model; underlined numbers are the best results across all models (same for Table~\ref{tab:results_mlqa_tydiqa}).}
\label{tab:results_xquad_filtered}
\end{table*}

\begin{table*}[t!]
\centering
\resizebox{\textwidth}{!}{
\begin{tabular}{lcccc|c|ccccc|c|}
& \multicolumn{5}{c}{MLQA} & & \multicolumn{5}{c}{TyDiQA-GoldP} \\
 \cmidrule{2-6} \cmidrule{8-12}
& en & de & es & hi & \bf avg & & en & fi & ko & ru & \bf avg \\
\midrule
mBERT\textsuperscript{*} & \multicolumn{1}{|c}{80.2 / 67.0} & 59.0 / 43.8 & \textbf{67.4} / 49.2 & 50.2 / \textbf{35.3} & 64.2 / 48.8 & \; & \multicolumn{1}{c}{\bf 75.3 / 63.6} & 59.7 / \textbf{45.3} & \bf 58.8 / 50.0 & 60.0 / 38.8 & \bf 63.5 / 49.4 \\
mBERT & \multicolumn{1}{|c}{\textbf{80.8} / 67.8} & 61.0 / 46.4 & 67.3 / 49.2 & 49.3 / 33.6 & 64.6 / 49.3 & & \multicolumn{1}{c}{74.3 / 61.8} & 60.3 / 44.0 & 57.3 / 46.7 & \textbf{62.5} / 42.3 & 63.6 / 48.7 \\
\; + ISDG & \multicolumn{1}{|c}{80.7 / \textbf{67.9}} & \bf 62.3 / 48.1 & 67.1 / \textbf{49.4} & \textbf{50.3} / 35.1 & \bf 65.1 / 50.2 & & \multicolumn{1}{c}{74.4 / 63.2} & \textbf{61.1} / 43.5 & 52.5 / 44.2 & 61.3 / \textbf{43.7} & 62.3 / 48.7 \\
\midrule
XLM-R\textsuperscript{*} & \multicolumn{1}{|c}{83.5 / 70.6} & 70.1 / 54.9 & 74.1 / \textbf{56.6} & 70.6 / 53.1 & 74.6 / 58.8 & & \multicolumn{1}{c}{71.5 / 56.8} & 70.5 / 53.2 & 31.9 / 10.9 & 67.0 / 42.1 & 60.2 / 40.8 \\
XLM-R & \multicolumn{1}{|c}{84.5 / 71.5} & 71.1 / 56.1 & 74.2 / 56.4 & 71.4 / 53.6 & 75.3 / 59.4 & & \multicolumn{1}{c}{73.6 / 61.3} & 74.2 / 58.2 & 59.4 / 47.8 & 69.5 / 46.8 & 69.2 / 53.5 \\
\; + ISDG & \multicolumn{1}{|c}{\bf \underline{84.9} / \underline{71.9}} & \bf \underline{71.2} / \underline{56.2} & \textbf{74.4} / 56.2 & \bf \underline{71.8} / \underline{54.0} & \bf \underline{75.6} / \underline{59.6} & & \multicolumn{1}{c}{\textbf{76.2} / \textbf{\underline{64.5}}} & \bf \underline{75.3} / \underline{59.4} & \bf 64.0 / 52.5 & \bf 70.7 / 51.2 & \bf 71.6 / 56.9 \\
\midrule
mT5\textsuperscript{*} & \multicolumn{1}{|c}{\textbf{\underline{84.9}}} / 70.7 & 68.9 / 51.8 & 73.5 / 54.1 & 66.9 / 47.7 & 73.6 / 56.1 & & \multicolumn{1}{c}{71.6 / 58.9} &  64.6 / 48.8 &  47.6 / 37.3 & 58.9 / 36.8 & 60.7 / 45.5 \\
mT5 & \multicolumn{1}{|c}{84.5 / 71.7} & 69.0 / 53.9 & 73.8 / 56.2 & 69.2 / 51.8 & 74.1 / 58.4 & & \multicolumn{1}{c}{73.3 / 60.9} & 71.5 / 54.5 & 60.8 / 51.1 & 68.1 / 44.8 & 68.4 / 52.8 \\
\; + ISDG & \multicolumn{1}{|c}{\bf \underline{84.9} / \underline{71.9}} & \bf 69.6 / 54.4 & \bf \underline{74.7} / \underline{56.7} & \bf 70.4 / 52.2 & \bf 74.9 / 58.8 & & \multicolumn{1}{c}{\bf \underline{76.3} / \underline{64.5}} & \bf 73.1 / 55.1 & \bf \underline{66.0} / \underline{56.5} & \bf \underline{73.3} / \underline{56.0} & \bf \underline{72.2} / \underline{58.0} \\
\end{tabular}}
\caption{MLQA results (left) and TyDiQA-GoldP results (right) (F1/EM) for each language.}
\label{tab:results_mlqa_tydiqa}
\end{table*}

We evaluate our models on three multilingual MRC benchmarks suggested by XTREME: XQuAD \citep{artetxe-etal-2020-cross}, MLQA \citep{lewis-etal-2020-mlqa}, TyDiQA-GoldP \citep{clark-etal-2020-tydi}. For XQuAD and MLQA, models are trained on English SQuAD v1.1 \citep{rajpurkar-etal-2016-squad} and evaluated directly on the test sets of each dataset in multiple target languages. For TyDiQA-GoldP, models are trained on its English training set and evaluated directly on its test sets. We use the evaluation scripts provided by XTREME, keeping the evaluation protocols identical. Standard metrics of F1 and exact-match (EM) are used.

As we use Stanza to obtain UD features, our experiments include languages that are supported by UD and also have similar prediction performance as the source language English, which largely keeps the obtained UD features to be consistent across languages.
Specifically, we compare the dependency parsing performance per language by the Labeled Attachment Score (LAS, the main evaluation metric for dependency parsing) provided by Stanza\footnote{\url{https://stanfordnlp.github.io/stanza/performance.html}}, and include any languages that currently have LAS score above 80. The resulting evaluation includes a total of 8 languages and 14 test sets in our experiments. With the active development of the UD project, more languages and higher feature quality are to be expected in the near future.

\subsection{Results}
\label{subsec:results}

The evaluation results for XQuAD are shown in Table~\ref{tab:results_xquad_filtered}, and the left and right part of Table~\ref{tab:results_mlqa_tydiqa} show the results for MLQA and TyDiQA-GoldP respectively.
In particular, mBERT\textsuperscript{*}, XLM-R\textsuperscript{*} and mT5\textsuperscript{*} denote the results reported from the original papers of XTREME and mT5; all other results are obtained from our re-implemented baselines and proposed models. Three different multilingual pretrained language models are experimented on all three datasets, and ``+ISDG'' shows the results of adding our ISDG encoder on the corresponding pretrained model.

The entire evaluation consists of 14 test sets in 8 languages. 
The best result for every test set, denoted by the underlined score of each column, is achieved by our ISDG encoder in terms of either F1 or EM.
The ISDG encoder also establishes the best on-average performance on all three datasets using either one of the three multilingual pretrained models, except for mBERT on TyDiQA-GoldP.
Specifically, the best on-average results of both XQuAD and MLQA are achieved by the ISDG encoder with XLM-R, while the encoder with mT5 shows the best results for TyDiQA-GoldP, improving upon its corresponding baseline by 3.8 F1 / 5.2 EM on average. On certain test sets, the improvement can be quite significant. For instance, ISDG brings 5.2 F1 / 11.2 EM improvement using mT5 on the TyDiQA-GoldP test set in Russian (ru).

The results per language indicate that although UD is designed to provide consistent features across languages, different languages do not benefit from the syntactic features equally, potentially due to the intrinsic differences among languages from the linguistic perspective, and the different feature quality across languages obtained from Stanza.
Nevertheless, most languages are indeed shown to have consistent performance boost. Some observations can be summarized as follows:
\begin{itemize}[noitemsep,nolistsep,leftmargin=*]
    \item English (en), German (de), Greek (el), Hindi (hi), Russian (ru), Finnish (fi) can get positive impact from UD features consistently on different datasets using either one of the pretrained models (improvement goes up to 5.2 F1).
    \item Spanish (es) gets positive impact from UD features overall; however, it can be dataset-specific, and does not outperform the baseline on XQuAD using mBERT or mT5.
    \item Korean (ko) gets significant improvement on TyDiQA-GoldP using XLM-R or mT5 (up to 5.2 F1 / 5.4 EM). However, the performance drops when using mBERT, likely because of the incompatibility between the wordpiece tokenizer of mBERT and Stanza tokenization on the segmentation of text in Korean.
\end{itemize}

Table~\ref{tab:results_mlqa_tydiqa} also shows that the improvement on TyDiQA-GoldP is higher than that on XQuAD and MLQA. For example, English (en) and Russian (ru) have 3 F1 and 5.2 F1 gain respectively on TyDiQA-GoldP when using ISDG encoder with mT5, which is much higher than the 0.9 F1 and 1.9 F1 gain on XQuAD under the same setting. As the training set for TyDiQA-GoldP is much smaller than SQuAD (the training set for XQuAD and MLQA), and only has 4.3\% size of the data as SQuAD, it suggests another potential advantage of utilizing UD features in the zero-shot setting. When the training data is not as enough on the source language, encoding universal syntactic features can help the model quickly learn the task objective and generalize to multiple languages.

\section{Analysis}
\label{sec:analysis}



\subsection{Ablation Study}
\label{subsec:ablation}

We first perform the ablation study of the ISDG encoder to examine the local and global graph encoding. We evaluate on the languages that have consistent performance boost on XQuAD to show the impact more explicitly. Table~\ref{tab:ablation} shows the result differences in F1 with three settings: only using POS features (skipping graph encoding entirely, similar to baselines but with UD tokenization and POS features), adding the local encoding component (+ L), adding both local and global components (+ L\&G).

\begin{table}[ht!]
\small
\centering
\resizebox{0.85\columnwidth}{!}{
\begin{tabular}{lccccc}
& en & de & el & hi & ru \\
\midrule
mBERT + POS & 83.9 & 71.8 & 63.8 & 58.3 & 71.7 \\
\; + L & +0.1 & +1.2 & +0.3 & +0.5 & +0.3 \\
\; + L\&G & +0.2 & +2.3 & +0.6 & +0.9 & +0.5 \\
\midrule
XLM-R + POS & 87.6 & 81.3 & 81.1 & 76.5 & 81.1 \\
\; + L & +0.6 & +0.5 & +0.4 & +0.2 & +0.2 \\
\; + L\&G & +1.0 & +0.8 & +0.8 & +0.4 & +0.2 \\
\midrule
mT5 + POS & 87.9 & 81.0 & 79.4 & 75.8 & 78.8 \\
\; + L & +0.5 & +0.8 & +0.7 & +0.6 & +0.8 \\
\; + L\&G & +0.8 & +1.5 & +1.1 & +1.1 & +1.7 \\
\end{tabular}}
\caption{Ablations on the ISDG encoder. Results (F1) are shown on XQuAD, collected from five runs on average. The improvement from the local and global components is largely consistent across the experimented languages.}
\label{tab:ablation}
\end{table}

The improvement from both components is consistent across the experimented languages, with the global encoding component contributing around 40\% of improvement on average, which shows the effectiveness of addressing the global multi-hop syntactic relations across sentences by encoding the approximated ``soft'' paths. Additionally, the model with only POS features can still have around 0.1 - 0.2 F1 improvement over the corresponding baseline, showing that the UD tokenization and POS features also contribute to the final performance trivially.

\subsection{Attentions on Global Encoding}
\label{subsec:attn-graph}

\begin{figure}[t]
\centering
\includegraphics[width=0.45\columnwidth]{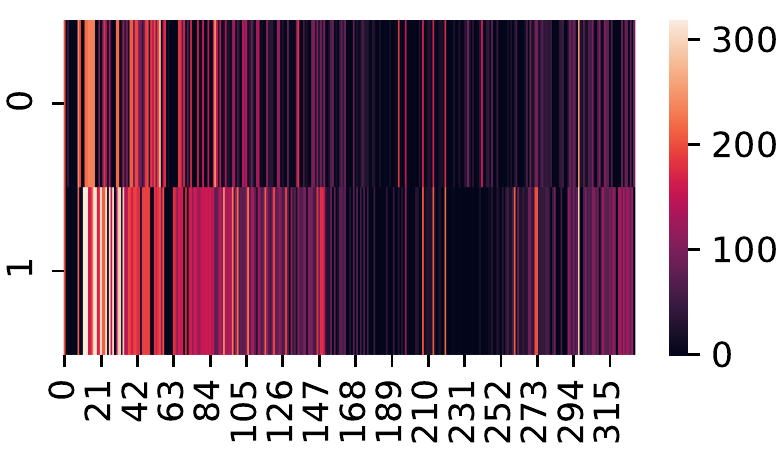}
\caption{The attention heat map of an input example in German, with XLM-R on XQuAD w.r.t attention distances. x-axis is the input sequence and y-axis represents the two attention heads. The distance is denoted by the temperature.}
\label{fig:dist}
\end{figure}

We next specifically look at the
attention distribution over the entire graph nodes in Eq~\eqref{eq:sp_sa}, to further understand how the global encoding brings improvement.
We keep track of the attentions at each attention head, and measure the attention distance of each node $i$, denoted by $D_i = |i - \text{argmax}_{j=1}^n \alpha^{\text{G}}_{ij}|$, which is the distance between its current position and the position to which it has the maximum attention weight. Figure~\ref{fig:dist} shows the heat map of an input example on two attention heads w.r.t the attention distance, with $D_i$ denoted by the temperature. Figure~\ref{fig:dist} suggests that it is fairly common for a graph node to have $D_i > 100$ (denoted by the high temperature), which means the node pays high attention to a likely cross-sentence node. It is especially common for nodes at the beginning of the sequence, as they are the nodes within the question, and heavily attend to the context.

In addition, we record the attentions and calculate the averaged attention distance using XLM-R on XQuAD. Our statistics show that it sits in the range of 50-60 and varies slightly by languages. By contrast, the vanilla self-attention in the last layer of pretrained model has averaged attention distance below 40. It shows that the attentions in the global component are good at modeling the long-term dependency, overcoming the drawback of the local component that only uses one-hop relations, and demonstrating the necessity to address global syntactic relations for stronger encoding of the input structure.
The attention distribution of the global encoding component also shows that the ``soft'' paths successfully activate cross-sentence information flow through the syntactic dependencies, albeit remaining an approximation of the true multi-hop paths.

\section{Conclusion}
\label{sec:conclusion}

In this work, we target to improve the direct zero-shot performance on the multilingual MRC task, by utilizing cross-linguistically consistent features from UD including POS and syntactic dependency relations.
Using the raw syntactic dependencies within each sentence, we build the ISDG to adapt to the multi-sentence input, and introduce the ISDG encoder to encode the obtained graph.
Especially, the encoder consists of both a local component that encodes one-hop relations, as well as a global component that encodes the global multi-hop relations by adopting the approximated ``soft'' paths between each node pair.
Experiments with three multilingual pretrained models on three datasets show that our ISDG encoder is able to improve zero-shot results consistently by a solid margin, up to 3.8 F1 / 5.2 EM improvement on average; around 40\% improvement is shown to come from the attentions on global syntactic encoding.

\bibliography{aaai22}

\end{document}